\documentclass[10pt, a4paper]{article}

\usepackage{lrec-coling2024} % this is the new style
\usepackage{times}
\usepackage{latexsym}
\usepackage{amsmath}
\usepackage{amsfonts}
\usepackage{comment}
\usepackage[T1]{fontenc}
\usepackage[utf8]{inputenc}

\usepackage{microtype}
\usepackage[ruled,vlined]{algorithm2e}
\usepackage{multirow}
\usepackage{amsmath, xparse}
\usepackage{makecell}
\usepackage{ragged2e}
\usepackage{xcolor}
\usepackage{graphicx}
\usepackage{subfigure}
\usepackage{booktabs,caption}
\usepackage[flushleft]{threeparttable}

\usepackage{inconsolata}

\title{\normalfont{  \textbf{Prompt-based Zero-shot Relation Extraction with Semantic
Knowledge Augmentation}}}

\name{Jiaying Gong, Hoda Eldardiry} 

\address{Virginia Tech\\
         Blacksburg, VA, USA\\
         \{gjiaying, hdardiry\}@vt.edu\\}

\abstract{
In relation triplet extraction (RTE), recognizing unseen relations for which there are no training instances is a challenging task.
Efforts have been made to recognize unseen relations based on question-answering models or relation descriptions.
However, these approaches miss the semantic information about connections between seen and unseen relations.
In this paper, We propose a prompt-based model with semantic knowledge augmentation (ZS-SKA) to recognize unseen relations under the zero-shot setting. We present a new word-level analogy-based sentence translation rule and generate augmented instances with unseen relations from instances with seen relations using that new rule. We design prompts with weighted virtual label construction based on an external knowledge graph to integrate semantic knowledge information learned from seen relations. 
Instead of using the actual label sets in the prompt template, we construct weighted virtual label words. We learn the representations of both seen and unseen relations with augmented instances and prompts. We then calculate the distance between the generated representations using prototypical networks to predict unseen relations.
Extensive experiments conducted on three public datasets FewRel, Wiki-ZSL, and NYT, show that ZS-SKA outperforms other methods under zero-shot setting. Results also demonstrate the effectiveness and robustness of ZS-SKA. 
 \\ \newline  }

\begin{document}

\maketitleabstract

\section{Introduction}
Relation triplet extraction (RTE) aims to extract both the pairs of entities and relations from unstructured text.
However, existing approaches based on supervised learning~\cite{ma-etal-2022-joint, zhong-chen-2021-frustratingly, ren-etal-2021-novel, huguet-cabot-navigli-2021-rebel-relation, zhu-etal-2019-graph, DBLP:journals/corr/abs-2004-03786} or few-shot learning~\cite{liu-etal-2022-simple, ijcai2022p407, han-etal-2021-exploring, Gao_Han_Liu_Sun_2019, ren-etal-2020-two, dong-etal-2020-meta} still require labeled data.
They can not catch up with a dynamic and open environment where new classes emerge.
In the real-world setting, the classes of instances are sometimes rare or never seen in the training data.
Thus, we tend to learn a model similar to the way humans learn and recognize new concepts.
Such a task is referred to as zero-shot learning (ZSL).
We follow the same definition of ZSL in~\cite{chia-etal-2022-relationprompt, chen2021zsbert} to conduct experiments.

\begin{figure}[htp] 
 \center{\includegraphics[height=4cm,width=7.5cm]{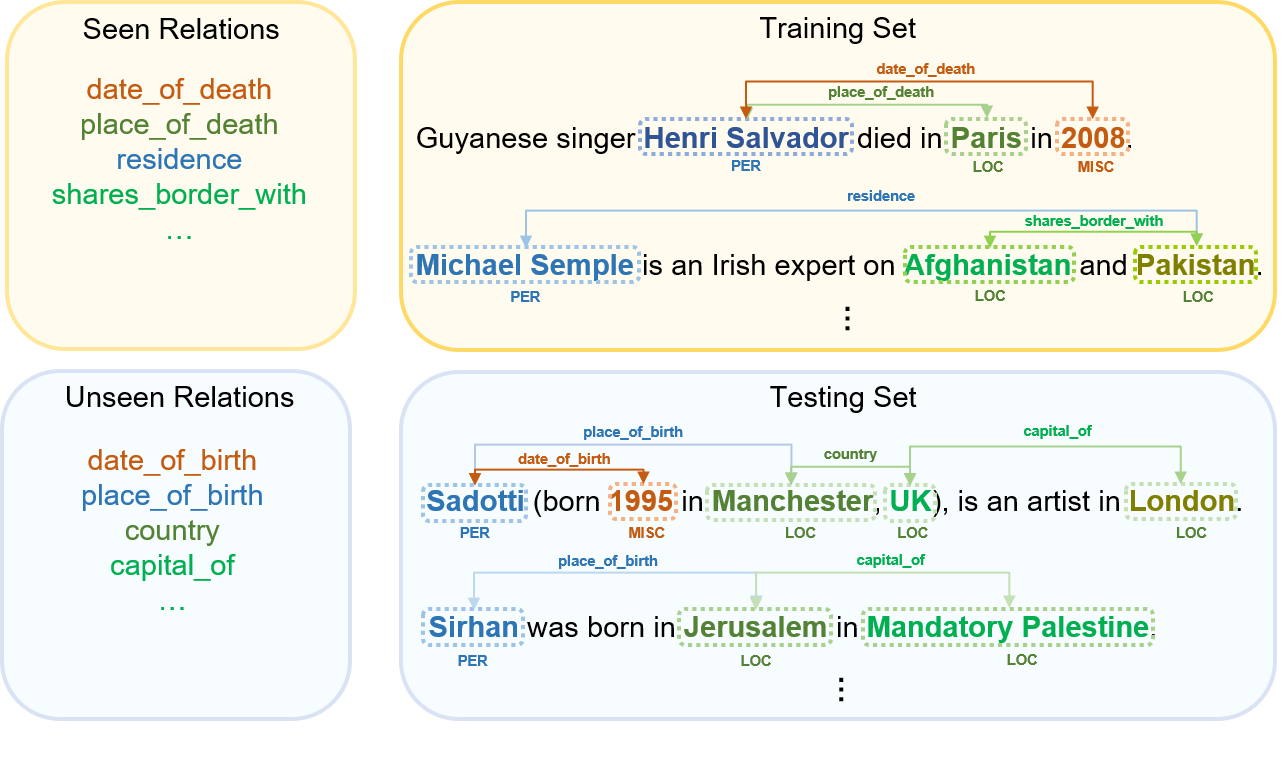}}
 \caption{\label{fig:example} Zero-shot RTE. There is no overlap of classes between training and testing data.} 
 \end{figure}

Zero-shot RTE aims to extract relation triplets in a sentence that is absent from the learning stage.
Figure~\ref{fig:example} shows an example of zero-shot RTE. 
The relation sets at the training and testing stages are disjoint.
The model for zero-shot RTE is only trained on the seen relations in the training stage and extracts triplets with unseen relations in the testing stage.
Existing approaches to zero-shot relation extraction still have limitations.
%Limitations
First, some models perform zero-shot relation extraction by question answering~\cite{levy-etal-2017-zero} or by using GPT-2 to help generate synthetic data~\cite{chia-etal-2022-relationprompt}.
These models have a strong assumption that an excellent additional deep learning model is learned, and that all values extracted from this model are correct. 
%This is impractical in the real-world setting.
Second, some existing studies formulate relation extraction as a text entailment task~\cite{obamuyide-vlachos-2018-zero}.
They only predict a binary label indicating whether the name entities in the given sentence can be described by a given description.
Third, some SOTA models leverage auxiliary information to tackle zero-shot tasks. They focus on class names/descriptions, losing the connection or relationships between seen relations and unseen relations~\cite{10.1145/3459637.3482403, chen2021zsbert, wang-etal-2022-rcl}.
Besides, these works mainly focus on zero-shot relation classification (ZSRC), which only predicts unseen relations instead of triplets in the format of <head entity, relation, tail entity>.
ZSRC has a strong assumption that two name entities are available for training.
However, it is not realistic that name entities are already provided.

%Inspired by Prompts and knowledge graph augmentation
To address the above challenges, we propose a prompt-based model with semantic knowledge augmentation (ZS-SKA) to perform zero-shot RTE.
%Data augmentation
We first implement data augmentation by a word-level sentence translation to generate augmented instances with unseen relations from training instances with seen relations.
%The super-class of the triplet <head entity, relation, tail entity> for augmented instances is the same as the triplet of training instances.
We follow a new generation rule introduced in Sec.~\ref{sec:data_augmentation} to generate high-quality augmented instances for training in zero-shot settings.
Note that ZS-SKA is trained only on labeled data from seen classes and augmented data generated from seen classes.

%Prompt
Secondly, inspired by prompt-tuning on pre-trained language models~\cite{DBLP:conf/eacl/SchickS21, DBLP:conf/naacl/SchickS21}, we design the prompts based on the knowledge graph to integrate semantic knowledge to generally infer the features of unseen relations using patterns learned from seen relations.
For the prompt design, we consider semantic knowledge information, including relation descriptions, super-class of relations and name entities, and a general knowledge graph to effectively learn the unseen relations.
%virtual label
Instead of using the real label word directly in the prompt template, we automatically search a set of appropriate label words based on the knowledge graph for each label.
The weight of each appropriate label word is calculated based on its semantic knowledge information in Sec.~\ref{sec:prompts}.
We calculate the distance between each appropriate label with the true label itself to help denoise the set of appropriate label words.
Then, we construct virtual label words in the prompt by weighted averaging all appropriate label word candidates.

\begin{figure*}[htp] 
 \center{\includegraphics[height=6cm,width=\textwidth]{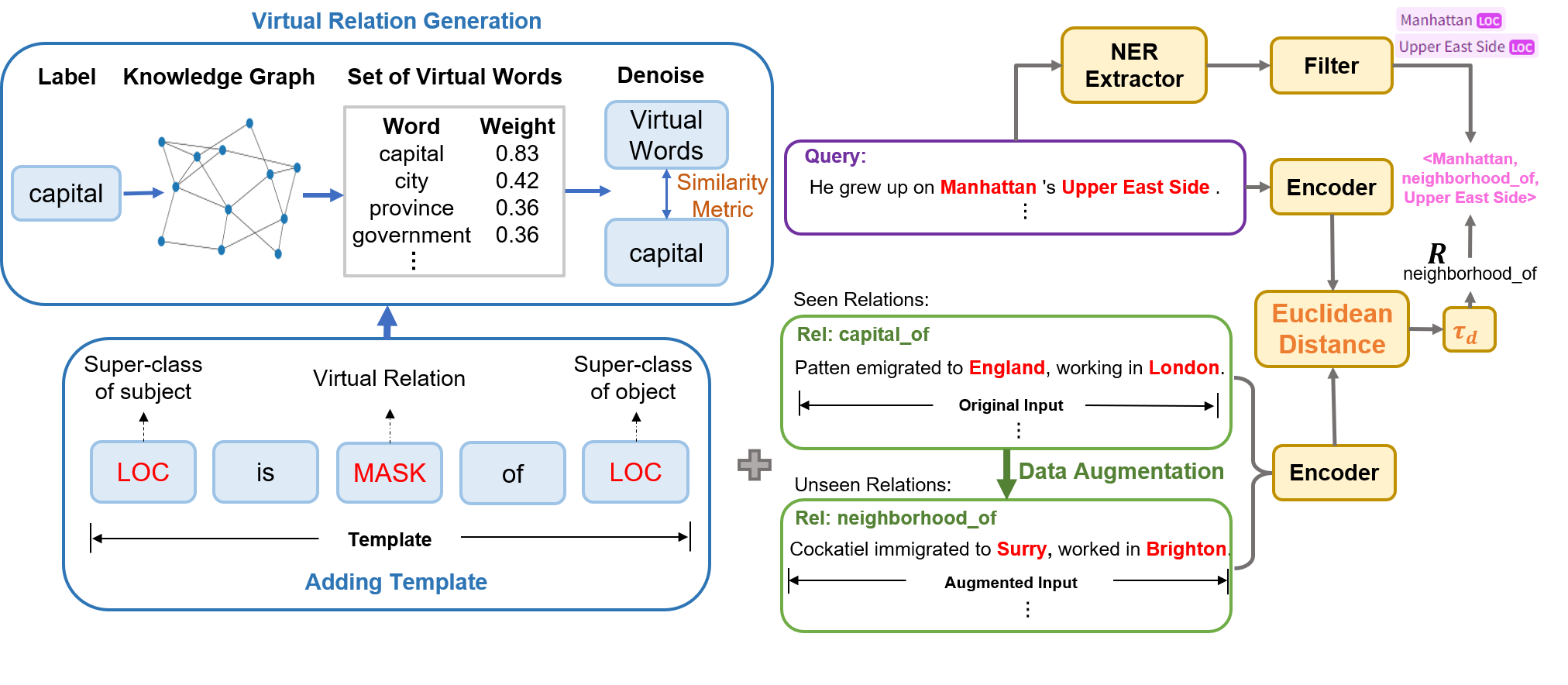}}
 \caption{\label{fig:framework} ZS-SKA overall architecture with components explained in Sec.~\ref{sec:3.2}.}
 \end{figure*}

Finally, we apply prototypical networks~\cite{NIPS2017_cb8da676} to compute a prototype representing each relation.
Each prototype is the mean vector of embedded and augmented sentences with prompts belonging to one relation.
Euclidean distance is calculated between query sentence embeddings with prototypes to predict relations.
%NER
A distance threshold explored in the validation set is applied to determine the number of relation triplets.
For name entity predictions, we use a name entity extractor to recognize different types of entities.
Then, sorted entity types are compared with the super-class of name entities in the prompt to determine the final relation triplets. The contributions:
\begin{itemize}
    \item We propose a prompt-based model with semantic knowledge augmentation (ZS-SKA) to extract triplets with unseen relations under the zero-shot setting.
    %enable zero-shot learning on relation classification tasks.
    Unlike some previous works, ZS-SKA considers semantic information from different granularities and does not rely on other large models with additional training.
	\item We present a new word-level sentence translation rule to generate augmented instances with unseen relations from instances with seen relations. The augmented sentences are then used as the training sets for unseen relations.
	%We construct augmented instances generated by a novel word-level translation rule and prompts with semantic knowledge built using the knowledge graph to learn representations for both seen and unseen relations. 
	\item We propose prompts for training based on an external knowledge graph to integrate semantic knowledge information learned from seen relations. We construct weighted virtual label words for the mask in the prompt template instead of using the actual label sets.
	\item We demonstrate that ZS-SKA significantly outperforms state-of-the-art methods for relation extraction with unseen relations under the ZSL setting on three public datasets. %Results show the effectiveness and robustness of ZS-SKA.
\end{itemize}

\section{Related Work}%Grammar Done
\subsection{Prompt Learning in NLP}
With the development of Generative Pre-trained Transformer 3 (GPT-3)~\cite{NEURIPS2020_1457c0d6}, prompt-based learning has received considerable attention.
Language prompts have been proven to be effective in downstream tasks leveraging pre-trained language models~\cite{DBLP:journals/corr/abs-1806-02847, davison-etal-2019-commonsense, petroni-etal-2019-language}.
Human-designed prompts have achieved promising results in few-shot learning for sentiment classification~\cite{DBLP:conf/eacl/SchickS21, DBLP:conf/naacl/SchickS21}.
To avoid labor-intensive prompt design, studies explore prompts that are generated automatically~\cite{DBLP:conf/emnlp/ShinRLWS20, DBLP:journals/tacl/JiangXAN20, gao2021making}.
However, most of the studies focus on supervised or few-shot learning on text classification~\cite{fei-etal-2022-beyond, hu2021knowledgeable, han2021ptr, gu2021ppt}, event detection~\cite{li-etal-2022-kipt}, relation classification~\cite{han2021ptr, DBLP:journals/corr/abs-2104-07650} and name entity recognition~\cite{li-etal-2022-prompt-based-text, huang-etal-2022-copner, ma2021templatefree}.
Inspired by these works, we explore prompt-based zero-shot learning in RTE.

\subsection{Zero-shot Relation Classification}
Relation classification is the problem of classifying relations given two name entities within a sentence.
Most existing works rely on sufficient human-labeled data or noisy labeled data by distant supervision.
When no training instances are available, some studies use zero-shot relation classification to extract unseen relations. 
This is typically done using question-answering models by listing questions that define the relation's slot values~\cite{levy-etal-2017-zero, cetoli-2020-exploring}.
Some studies formulate relation extraction as a text entailment task~\cite{obamuyide-vlachos-2018-zero}.
Some studies utilize the accessibility of the relation descriptions to get the information for unseen relations~\cite{obamuyide-vlachos-2018-zero, Qin_Wang_Chen_Zhang_Xu_Wang_2020, 10.1145/3459637.3482403, chen2021zsbert, sainz-etal-2021-label, liu-etal-2022-pre}.
However, these models only utilize class names semantic information, losing the connections between relations.
%Similarity is then calculated between the embedding of testing instances and the embedding of label information to extract unseen relations.
Other studies focus on establishing the connection between relations with knowledge graph~\cite{li-etal-2020-logic} or contrastive learning~\cite{wang-etal-2022-rcl}.
Nevertheless, all these works only focus on relation classification instead of relation triplet extraction.
In the real world, it is not practical and realistic that two name entities are provided for training.
Therefore, we focus on a more complex and realistic task: extracting both name entities and their relations.

\subsection{Zero-shot Relation Triplet Extraction}
%Relation triplet extraction (RTE) aims at extracting both head and tail entities, and their corresponding relations.
%RTE in the zero-shot setting means that there's no overlap of relations (classes) between the training set and the testing set.
Current approaches focusing on zero-shot relation triplet extraction (RTE) require large computing resources and additional deep-learning models.
For example, RelationPrompt requires to fine-tune a pre-trained GPT-2 (124M parameters) as the relation generator~\cite{chia-etal-2022-relationprompt}.
PCRED needs to train the entity boundary detection module by four neural network layers to get the possible boundaries for name entities in the triplet~\cite{lan2022pcred}.
ZETT views relation extraction as a template-filling problem, fine-tuning T5 to get the ranking score for potential triplets~\cite{kim2022zero}.
However, ZETT can not discriminate against similar relations because the connections of different relations are lost.
Inspired by data augmentation from knowledge graph in text classification~\cite{zhangkumjornZeroShot, 9466939} and prompt-based few-shot learning~\cite{hu2021knowledgeable}, we propose a prompt-based zero-shot RTE framework (ZS-SKA) incorporating external knowledge from the knowledge graph.
Different from these existing works, the data augmentation module and name entity recognition module in ZS-SKA do not require any additional training.
ZS-SKA can better catch the connections between relations due to the incorporated knowledge graph in the prompt template.

\section{Methodology}%Grammar Done
In this section, we introduce the overall framework as shown in Figure~\ref{fig:framework} of ZS-SKA.

\subsection{Problem Definition}%Done
%Zero-shot Definition.
To do zero-shot relation extraction, we adopt the problem setting in~\cite{chia-etal-2022-relationprompt, 10.1145/3459637.3482403} for zero-shot relation triplet extraction and setting in~\cite{chen2021zsbert} for zero-shot relation classification. 
We also conduct ablation experiments following the zero-shot definition in~\cite{yinroth2019zeroshot} which is a generalized zero-shot setting where partial labels are unseen. 
%Given a set $Se$ of seen classes and a set $U$ of unseen class, and labeled instances for $Se$, we learn a model $M:X\rightarrow Y=Se \cup U$.
%A general formulation for zero-shot is defined as:
Given labeled instances belonging to a set of seen classes $S$, a model $M:X\rightarrow Y$ is learned, where $Y=Se \cup U$; $U$ is the unseen class.
%Zero-shot relation triplet extraction definition
For zero-shot RTE, let $R_{s}=\begin{Bmatrix} r_{s}^{1},\cdots,r_{s}^{m} \end{Bmatrix}$ and $R_{u}=\begin{Bmatrix} r_{u}^{1},\cdots,r_{u}^{n} \end{Bmatrix}$ denote the sets of seen and unseen relations, where $m=|R_{s}|$ and $n=|R_{u}|$ are the number of relations in the two disjoint sets, i.e., $R_{s}\cap R_{u} = \emptyset$.
The set of relations $R$ is pre-defined.
The input of the training set consists of (1) seen relations $R_{s}$ with input sentences $X_{i}$, (2) unseen relations $R_{u}$ with super-class $S(e_{i_{h}})$, $S(e_{i_{t}})$ super-class of two name entities, and (3) external knowledge graph $G$. Here super-class is the hypernym of the item. For example, location (LOC) is the super-class of New York. 
The output of the model is a triplet in the format of <head ($e_{i_{h}}$), relation (r), tail ($e_{i_{t}}$)> or a set of triplets if $X_{i}$ contains multiple triplets.
Our goal is to train a zero-shot relation triplet extraction model $M$ to (1) learn the representations of both seen and unseen relations, (2) predict new triplet <$e_{i_{h}}$, $r_{u}$, $e_{i_{t}}$>, where the relation $r_{u}$ is not seen during the training phase.
$M$ is learned by minimizing the semantic distance between the embedding of the input and relation representations built from the knowledge graph $G$.
%Zero-shot relation classification definition
For the zero-shot RC, the only difference with zero-shot RTE is that the two name entities $e_{i_{h}}$ and $e_{i_{t}}$ are known information with the input sentence $X_{i}$. 
Therefore, the model only needs to predict $r_{u}$ given $X_{i}$ with $e_{i_{h}}$ and $e_{i_{t}}$.

\subsection{Semantic Knowledge Augmentation}~\label{sec:3.2}
\vspace{-8mm}
\subsubsection{Data Augmentation}~\label{sec:data_augmentation}%Done
To enable the model to detect unseen relations without labeled training instances, we first do data augmentation by translating a sentence from its original seen relation to a new unseen relation using an analogy.
In the word level, we adopt 3CosMul~\cite{levy-goldberg-2014-linguistic}, where we use the top 10 similar words to return, to get the candidates of new words $w_{u}$:
\begin{equation}
    w_{u}=\underset{x\in V}{argmax}\frac{cos(x, r_{u})\cdot cos(x,w_{s})}{cos(x,r_{s})+ \epsilon }
\end{equation}
where $V$ is the vocabulary set, $cos(\cdot)$ is the cosine similarity, $r_{u}$ is the unseen relation, $r_{s}$ is the seen relation, $w_{s}$ is the word in seen class and $\epsilon$ is a small number to prevent division by zero.

\begin{algorithm}[!htbp]
\small
\SetKwInOut{KIN}{Input}
\SetKwInOut{KOUT}{Output}
\caption{Sentence Generation for Unseen Relations}
\label{alg:data} 
\KIN{sentence $x_{i}=[w_1^i,\cdots,w_n^i]$, two name entities $e_{i_{h}}$ and $e_{i_{t}}$, original 
relation label sets $R_s$, target unseen relation label $r_{u}$}
\KOUT{sentence $x_{i}^{u}$ with relation $r_{u}$}
%replace\_dict = $dict()$; $x_{i}^{u}=[]$\\
\For{$r_{s} \in R_{s}$}
{
    \eIf {$S(r_{u}) == S(r_{s})$ and $S(e_{u}) == S(e_{s})$}
    {
    \For{$w \in x_{i}$}
    {
    \eIf {is\_valid\_pos($w$)}
    {
    $w_{u} = 3CosMul(w, r_{u}, r_{s})$  \\
    $x_{i}^{u}$.append($w_{u}$)
    }{$x_{i}^{u}$.append($w$)}
    }
    }
    {Continue}
}
return $x_{i}^{u}$
\end{algorithm}

At the sentence level, we follow Algorithm~\ref{alg:data} to translate a sentence of relation $r_{s}$ into a new sentence of relation $r_{u}$.
To be more specific, we translate all nouns, verbs, adjectives, and adverbs in the seen sentence to a new sentence.
We do the translation when the super-class of $r_{s}$ and the super-class of two corresponding name entities in $r_{s}$ are the same as the super-class of $r_{u}$ and the super-class of two related name entities in $r_{u}$.
If the number of $r_{s}$ that conforms to the rules is larger than one, we take all the translated sentences and randomly select the same number as other seen relations to make a balanced training set.

\subsubsection{Prompts from Knowledge Graph}~\label{sec:prompts}%Done
For relation extraction, the core issue is to extract the possible triplets from all aspects and granularities.
For zero-shot tasks, we design prompts used as training instances to help train the model because there is no real training data available.
We construct prompts based on the external knowledge graph ConceptNet~\cite{Speer2013}, a knowledge graph that connects words and phrases of natural language with labeled edges, for zero-shot relation extraction.
Nodes in ConceptNet are entities, and edges connecting two nodes are semantic relations between the entities.
Because of the relation extraction task, we wrap the input sequence with a \textit{template}, which is a piece of natural language text.
To be more specific, we build prompts as `$S(e_{i_{h}})$ is [MASK] of $S(e_{i_{t}})$'. We consider different locations of prompts such as before and after the input sentence. There is a similar performance, so we put the prompts after each input sentence.
The [MASK] here is a virtual label word $r_{v}$ representing the relation between $S(e_{i_{h}})$ and $S(e_{i_{t}})$.
Unlike using real words, we build the virtual label word that can primarily represent the relation in each sentence.
Instead of building a virtual label word by simply using the mean vector of the top\_k high-frequency words~\cite{ma2021templatefree}, we build our virtual label word based on a knowledge graph using the following strategy.

\begin{algorithm}[!htbp]
\small
\SetKwInOut{KIN}{Input}
\SetKwInOut{KOUT}{Output}
\caption{Virtual Label Generation}
\label{alg:prompts} 
\KIN{word $w_{i}$, relation $r_{c}$, threshold $\tau_{s}$, number of hop $K$, Knowledge Graph $G$, number of virtual label $n$}
\KOUT{virtual label $r_{v}$}
\For{$w_{i} \in V$}
{
    \eIf {$\frac{w_{i}\cdot r_{c} }{|w_{i}|\times |r_{c}|}\geq \tau_{s} $}
    {
    $v_{1}$ = 0, $v_{2}$, $v_{3}$, $v_{ave}$ = []\\
    \eIf {$w_{i} \in G$}
    {
    $v_{1}$ = 1
    }{$v_{1}$= 0}
    \For{$k \in K$}
    {
    hops = find\_neighbors($w_{i}$) $\in$ $G$\\
    \eIf {hops}
    {
    $v_{2}$.append(any(hops))\\
    $v_{3}$.append(sum(hops))\\
    $v_{ave}$.append(mean(hops))
    }{$v_{2}$,$v_{3}$,$v_{ave}$.append(0)}
    $\alpha _{w_{i}} = \frac{\sum v}{Dim(v)}$
    }
    }
    {Continue}
    $\gamma _{v} = \frac{\alpha _{w_{i}}\cdot E(w_{i})+\cdots +\alpha _{w_{n}}\cdot E(w_{n})}{\sum \alpha }$
}
return $r_{v}$
\end{algorithm}

We firstly represent a relation $r$ as five sets of nodes in ConceptNet by processing the class label $r_{c}$, class hierarchy $S(r_{c})$, class description $D(r_{c})$ and hierarchy of two name entities $S(e_{i_{h}})$ and $S(e_{i_{t}})$.
We consider whether a word $w_{i}$ is related to the members of the five sets above within $K$ hops or not.
The value of $K$ is determined through the grid search on the validation set.
For each of the five sets above, we consider $v_{1}$ (whether $w_{i}$ is a node in $G$ in that set), $v_{2}$ (whether $w_{i}$'s neighbor is a node in $G$), $v_{3}$ (number of neighbors of $w_{i}$ in $G$). 
% \begin{itemize}
%     \item $v_{1}$: whether $w_{i}$ is a node in $G$ in that set.
%     \item $v_{2}$: whether $w_{i}$'s neighbor is a node in $G$.
%     \item $v_{3}$: number of neighbors of $w_{i}$ in $G$.
% \end{itemize}
The above values associated with each set demonstrate the semantic distance of $w_{i}$ and the corresponding set.
The detailed construction of virtual label $r_{v}$ is shown in Algorithm~\ref{alg:prompts}.

\subsection{Model Architecture and Training}%Done
\subsubsection{Instance Encoder}%Done
% \begin{figure}[htp] 
%  \center{\includegraphics[height=6cm,width=7.5cm]{Encoder.png}}
%  \caption{\label{fig:encoder} BERT-CNN Instance Encoder.} 
%  \end{figure}
%Figure~\ref{fig:encoder} shows the architecture of the encoder used in this paper.
We first tokenize and lemmatize all words in a sentence.
Two special tokens [CLS] and [SEP] are appended to the first and last positions, respectively.
Then BERT~\cite{devlin-etal-2019-bert} is used to generate the contextual representation for each token $w_{i}$.
Because relation is not only related to the original name entities in augmented sentences generated by data augmentation, we have not used any position embeddings to show the positions of $e_{i_{h}}$ and $e_{i_{t}}$.
Let $h_{i}$ represent the hidden state of the input sentence.
We use $CNN(\cdot)$ ReLU and a max-pooling layer $max(\cdot)$, to derive the representation:
\begin{equation}
    h_{i} = max(ReLU(CNN(x_{i})))
\end{equation}
where $x_{i}$ is the tokenized input sentence:
\begin{equation}
    x_{i} = w_{i-\frac{n-1}{2}},\cdots ,w_{i+\frac{n-1}{2}}
\end{equation}
We obtain the hidden state vectors of prompts $h_{p}$:
\begin{equation}
    h_{p} = E(S(e_{i_{h}})) \oplus E(r_{v}) \oplus E(S(e_{i_{t}}))
\end{equation}
where $E(\cdot)$ is the embedding function, $S(\cdot)$ is the super-class of the input word and $r_{v}$ is the virtual label embedding.
The final representation for each instance is the concatenation of $h_{i}$ and $h_{p}$.

\subsubsection{Name Entity Extractor}~\label{sec:ner}
ZS-SKA includes a name entity recognition encoder to extract name entities given the input $X_{i}$ with predicted relation $r_{i}$.
A fine-tuned BERT~\footnote{https://huggingface.co/dslim/bert-base-NER} is implemented to recognize different types of entities.
Each entity $e_{i}$ will be assigned a possibility score $score_{i}$ with entity types $T(e_{i})$ (i.e. B-PER, I-LOC).
\begin{equation}
<e_{i}, T(e_{i}), score_{i}> = E_{NER}(X_{i})
\end{equation}
where $E_{NER}$ is the name entity extractor based on BERT.
Based on the predicted relation sets $R_{i}$, super-classes of the possible name entities $S(e_{i_{h}})$ and $S(e_{i_{t}})$ can be accessed from the prompt templates.
Then, we filter out the entities whose types are different from the super-classes in the prompt template for each relation:
\begin{equation}~\label{equ:e}
E_{i} = \left \{ e_{i} | T(e_{i}) = S(e_{i}), score_{i} >= \tau_{e} \right \}
\end{equation}
where $E_{i}$ is the possible entity sets after the filter, $S(e_{i})$ is the super-class of the target relation in the prompt template and $\tau_{e}$ is the threshold for the possibility of name entity types.

\subsubsection{Model Training}~\label{sec:threshold}%Done
The objective of training ZS-SKA is to minimize the distance between each instance embedding $h_{i}\oplus h_{p}$ and the prototype $c_{i}$ embedding representing each learned relation (different colors in prototypes representation in Figure~\ref{fig:framework}).
Instead of using a softmax layer to classify seen relations and unseen relations, we adopt prototypical networks to compute a prototype for each relation after the BERT-CNN encoder.
Each prototype is the average instance embeddings belonging to one relation:
\begin{equation}
    c_{i}=\frac{1}{N}\sum_{i=1}^{N}f_{\phi }(h_{i}\oplus h_{p})
\end{equation}
where $c_{i}$ represents the prototype for each relation, $f_{\phi }$ is the BERT-CNN encoder, $h_{i}$ is the representation for each original or augmented sentence and $p_{i}$ is denotes the prompt embeddings introduced in Sec.~\ref{sec:prompts}.
The probabilities of the relations in $R_{s}$ and $R_{u}$ for a query instance $x$ is calculated as:
\begin{equation}
    p_{\phi }(y=r_{i}|x)=\frac{exp(-d(f_{\phi }(h_{i}\oplus h_{p}),c_{i}))}{\sum_{j=1}^{|R|}exp(-d(f_{\phi }(h_{i}\oplus h_{p}),c_{j}))}
\end{equation}
where $d(.)$ is the Euclidean distance for two vectors.
For multiple zero-shot RTE, we set a distance threshold $\tau_{d}$ to determine the number of possible unseen relations.
During the inference phase, ZS-SKA predicts the relation set $R_{i}$ by comparing the normalized distance $d(x_{i})$ with the threshold $\tau_{d}$:
\begin{equation}~\label{equ:r}
R_{i} =\left \{ r_{i} | softmax(d(x_{i}))< \tau_{d} , r_{i}\in R \right \}
\end{equation}
where $d(.)$ is the Euclidean distance. The final distance threshold $\tau_{d}$ for the testing phase is chosen by the threshold value that has the best performance in the evaluation phase.
For zero-shot RTE, the final relation triplets are the combination of $E_{i}$ from Equ.~\ref{equ:e} and $R_{i}$ from Equ.~\ref{equ:r}.
For zero-shot RC, only $R_{i}$ is provided for the result.

\begin{table*}[]
\small
\caption{Results for Zero-Shot Relation Triplet Extraction.}
\label{tab:triplet}
\centering
\tabcolsep=0.2cm
\begin{tabular}{clcccccccc}
\hline
          &                                             & \multicolumn{4}{c}{FewRel}                                                             & \multicolumn{4}{c}{Wiki\_ZSL}                                     \\
\#unseen  &                                             & Single         & \multicolumn{3}{c}{Multi}                                             & Single         & \multicolumn{3}{c}{Multi}                        \\
relations & \multicolumn{1}{c}{Model}                   & Acc.           & Pre.           & Rec.           & F1                            & Acc.           & Pre.           & Rec.           & F1       \\ \hline
          & \multicolumn{1}{l|}{TabSeq~\cite{wang-lu-2020-two}}          & 11.82          & 15.23          & 1.91           & \multicolumn{1}{c|}{3.40}           & 14.47          & 43.68          & 3.51           & 6.29           \\
          & \multicolumn{1}{l|}{NoGen~\cite{chia-etal-2022-relationprompt}}                  & 11.49          & 9.45           & \textbf{36.74} & \multicolumn{1}{c|}{14.57}          & 9.05           & 15.58          & \textbf{43.23} & 22.26          \\
          & \multicolumn{1}{l|}{RelPrompt~\cite{chia-etal-2022-relationprompt}}         & 22.27          & 20.80          & 24.32          & \multicolumn{1}{c|}{22.34}          & 16.64          & 29.11          & 31.00          & 30.01          \\
m=5       & \multicolumn{1}{l|}{ZETT\textsubscript{T5-small}~\cite{kim2022zero}}     & 26.34          & 31.12          & 30.01          & \multicolumn{1}{c|}{30.53}          & 20.24          & 31.62          & 32.41          & 31.74          \\
          & \multicolumn{1}{l|}{ZETT\textsubscript{T5-base}~\cite{kim2022zero}}      & 30.71          & 38.14          & 30.58          & \multicolumn{1}{c|}{33.71}          & 21.49          & 35.89          & 28.38          & 31.74          \\
          & \multicolumn{1}{l|}{PCRED~\cite{lan2022pcred}}                  & 22.67          & 43.91          & 34.97          & \multicolumn{1}{c|}{\textbf{38.93}} & 18.40          & 38.14          & 36.84          & 37.48          \\
          & \multicolumn{1}{l|}{\textbf{ZS-SKA (ours)}} & \textbf{32.86} & \textbf{57.50} & 26.24          & \multicolumn{1}{c|}{36.04}          & \textbf{44.00} & \textbf{66.70} & 27.24          & \textbf{38.68} \\ \hline
          & \multicolumn{1}{l|}{TabSeq~\cite{wang-lu-2020-two}}          & 12.54          & 28.93          & 3.60           & \multicolumn{1}{c|}{6.37}           & 9.61           & 45.31          & 3.57           & 6.4            \\
          & \multicolumn{1}{l|}{NoGen~\cite{chia-etal-2022-relationprompt}}                  & 12.40          & 6.40           & \textbf{41.70} & \multicolumn{1}{c|}{11.02}          & 7.10           & 9.63           & \textbf{45.01} & 15.70          \\
          & \multicolumn{1}{l|}{RelPrompt~\cite{chia-etal-2022-relationprompt}}         & 23.18          & 21.59          & 28.68          & \multicolumn{1}{c|}{24.61}          & 16.48          & 30.20          & 32.31          & 31.19          \\
m=10      & \multicolumn{1}{l|}{ZETT\textsubscript{T5-small}~\cite{kim2022zero}}     & 23.07          & 25.52          & 29.61          & \multicolumn{1}{c|}{27.28}          & 14.37          & 19.86          & 27.71          & 22.83          \\
          & \multicolumn{1}{l|}{ZETT\textsubscript{T5-base}~\cite{kim2022zero}}      & 27.79          & 30.65          & 32.44          & \multicolumn{1}{c|}{31.28}          & 17.16          & 24.49          & 26.99          & 24.87          \\
          & \multicolumn{1}{l|}{PCRED~\cite{lan2022pcred}}                  & 24.91          & 30.89          & 29.90          & \multicolumn{1}{c|}{30.39}          & 22.30          & 27.09          & 39.09          & 32.00          \\
          & \multicolumn{1}{l|}{\textbf{ZS-SKA (ours)}} & \textbf{34.03} & \textbf{60.48} & 23.22          & \multicolumn{1}{c|}{\textbf{33.28}} & \textbf{26.40} & \textbf{45.38} & 29.27          & \textbf{35.30} \\ \hline
          & \multicolumn{1}{l|}{TabSeq~\cite{wang-lu-2020-two}}          & 11.65          & 19.03          & 1.99           & \multicolumn{1}{c|}{3.48}           & 9.20           & \textbf{44.43} & 3.53           & 6.39           \\
          & \multicolumn{1}{l|}{NoGen~\cite{chia-etal-2022-relationprompt}}                  & 10.93          & 4.61           & \textbf{36.39} & \multicolumn{1}{c|}{8.15}           & 6.61           & 7.25           & \textbf{44.68} & 12.34          \\
          & \multicolumn{1}{l|}{RelPrompt~\cite{chia-etal-2022-relationprompt}}         & 18.97          & 17.73          & 23.20          & \multicolumn{1}{c|}{20.08}          & 16.16          & 26.19          & 32.12          & 28.85          \\
m=15      & \multicolumn{1}{l|}{ZETT\textsubscript{T5-small}~\cite{kim2022zero}}     & 21.08          & 16.20          & 23.22          & \multicolumn{1}{c|}{18.90}          & 10.74          & 14.96          & 19.31          & 16.79          \\
          & \multicolumn{1}{l|}{ZETT\textsubscript{T5-base}~\cite{kim2022zero}}      & \textbf{26.17} & 22.50          & 27.09          & \multicolumn{1}{c|}{24.39}          & 12.78          & 19.45          & 23.31          & 21.21          \\
          & \multicolumn{1}{l|}{PCRED~\cite{lan2022pcred}}                  & 25.14          & 27.00          & 23.55          & \multicolumn{1}{c|}{25.16}          & \textbf{21.64} & 25.37          & 33.80          & 28.98          \\
          & \multicolumn{1}{l|}{\textbf{ZS-SKA (ours)}} & 23.86          & \textbf{37.29} & 19.13          & \multicolumn{1}{c|}{\textbf{25.29}} & 20.26          & 31.23          & 27.20          & \textbf{29.19} \\ \hline
\end{tabular}
\end{table*}

\section{Experiments}
We conduct several experiments with ablation studies on three public datasets: FewRel, Wiki-ZSL, and NYT to show that our proposed model outperforms other existing state-of-the-art models, and our proposed model is more robust compared with the other models in the zero-shot learning tasks. 
\subsection{Evaluation Settings}
\subsubsection{Dataset}%Done
In our experiments, we evaluate our model~\footnote{Code: \url{https://github.com/gjiaying/ZS-SKA}} over three widely used datasets: FewRel~\cite{han-etal-2018-fewrel}, Wiki-ZSL~\cite{chen2021zsbert} and NYT~\cite{10.1007/978-3-642-15939-8_10}. 
FewRel and Wiki-ZSL are two balanced datasets and NYT is an unbalanced dataset.
The statistics of FewRel, Wiki-ZSL, and NYT datasets are shown in Table~\ref{tab:dataset}.
We provide a more detailed description in Sec.~\ref{sec:app_data}.

\subsubsection{Baselines and Evaluation Metrics}%Done
For RTE in ZSL, we compare our proposed model ZS-SKA with six SOTA zero-shot RTE models: \textbf{TableSequence}~\cite{wang-lu-2020-two}, \textbf{NoGen}~\cite{chia-etal-2022-relationprompt}, \textbf{RelationPrompt}~\cite{chia-etal-2022-relationprompt}, \textbf{ZETT}~\cite{kim2022zero} with two sizes of T5 models (T5-small and T5-base), and \textbf{PCRED}~\cite{lan2022pcred}.
For the zero-shot RC task, We compare our proposed model to eight existing RC models on all three public datasets to evaluate the model's ability to detect unseen relations.
For clean FewRel and Wiki-ZSL datasets, we compare our model with \textbf{CNN}~\cite{zeng-etal-2014-relation}, \textbf{Bi-LSTM}~\cite{zhang-etal-2015-bidirectional}, \textbf{Attentional Bi-LSTM}~\cite{zhou-etal-2016-attention}, \textbf{R-BERT}~\cite{10.1145/3357384.3358119}, \textbf{ESIM}~\cite{Chen-Qian:2017:ACL}, \textbf{CIM}~\cite{DBLP:journals/corr/RocktaschelGHKB15}, \textbf{ZS-BERT}~\cite{chen2021zsbert}, and \textbf{NoGen}~\cite{chia-etal-2022-relationprompt}.
The eight baselines above are reported by~\cite{chen2021zsbert} and ~\cite{chia-etal-2022-relationprompt}.
We also compare the robustness of our model with the most SOTA re-implemented \textbf{ZS-BERT}, \textbf{NoGen} and \textbf{RelationPrompt}.
For noisy NYT dataset, we compare our model with the re-implemented \textbf{CDNN}~\cite{zeng-etal-2014-relation}, \textbf{REDN}~\cite{DBLP:journals/corr/abs-2004-03786} and \textbf{ZSLRC}~\cite{10.1145/3459637.3482403}.
The evaluation metric for a single RTE is Accuracy (Acc.) because each sentence only includes one gold triplet. The evaluation metrics for multiple RTE are Precision (Pre.), Recall (Rec.), and F1-score (F1), because there are at least two gold triplets in the testing set.
For RC evaluation, we also use Precision, Recall, and F1-score, similar to those used for the above baselines.

\begin{table*}[]
\small
\caption{RC results with different $m$ values on NYT.}
\label{tab:mainresultun}
\centering
\begin{tabular}{lcccccc}
\hline
                & \multicolumn{3}{c}{m=15}                         & \multicolumn{3}{c}{m=30}   \\
                & Precision      & Recall         & F1             & Precision & Recall & F1    \\ \hline
CDNN        & 27.94 $\pm$ 0.52         & 44.10     $\pm$ 0.44      & 33.72  $\pm$ 0.46      & 10.17 $\pm$ 0.86    & 25.62 $\pm$ 0.71  & 14.23 $\pm$ 0.77\\
REDN           & 66.52  $\pm$ 0.47        & 65.47     $\pm$ 0.62     & 66.98     $\pm$ 0.55     & 57.19 $\pm$ 0.60    & 56.80  $\pm$ 0.70 & 56.99 $\pm$ 0.62 \\
ZSLRC          & 96.06 $\pm$ 0.35         & 93.84  $\pm$ 0.21        & 93.59    $\pm$ 0.31      & 94.81 $\pm$ 0.38     & \textbf{90.46$\pm$0.22}   & 89.76  $\pm$ 0.29\\ \hline
\textbf{ZS-SKA} & \textbf{96.23 $\pm$ 0.08} & \textbf{94.68 $\pm$ 0.12} & \textbf{94.42 $\pm$ 0.11} & \textbf{95.91 $\pm$ 0.20}     & 90.38 $\pm$ 0.32   & \textbf{91.27 $\pm$ 0.28}  \\ \hline
\end{tabular}
\end{table*}

\subsubsection{Parameter Settings}%Done
We follow the experiment settings as ~\cite{chia-etal-2022-relationprompt} and ~\cite{chen2021zsbert} to enable zero-shot RTE and zero-shot RC tasks.
We randomly select $m$ unseen relations and remove all the instances related to these $m$ relations in the training set to ensure that these $m$ relations have not appeared in training data. 
$m$ is varied to examine how performance is affected.
More details of hyperparameter settings are discussed in Sec.~\ref{sec:app_para} and Table~\ref{tab:parameter-3}.

\subsection{Results and Discussion}
\subsubsection{Zero-shot Relation Triplet Extraction}

\paragraph{Main Results}
Table~\ref{tab:triplet} shows the results of both single and multiple RTE on FewRel and Wiki\_ZSL in ZSL.
The results of our proposed model are reported by the average of five runs.
%All baselines results are reported by their original papers.
For single RTE, we observe that ZS-SKA significantly outperforms other baselines when m=5 and m=10. 
%Reason
%We conjecture that the reason for the decrease in accuracy of ZS-SKA when m=15 is because as the number of unseen relations increases, the number of seen relations decreases.
%This results in a less diverse set of augmented data based on Algorithm~\ref{alg:data}.
From Table~\ref{tab:triplet}, ZS-SKA demonstrates better performance on Wiki\_ZSL than on FewRel, as it shows a greater increase in accuracy compared to the strongest baseline.
This indicates that our proposed model is more robust and effective on the dataset with more classes as Wiki\_ZSL has 113 relations and FewRel has 80 relations in all.
For multiple RTE, ZS-SKA consistently achieves the best F1 score in all settings, except when m=5 on the FewRel dataset. Compared to other baselines, ZS-SKA achieves a relatively high precision score, resulting in a better F1 score. Downstream applications for RTE, such as building knowledge graphs using the extracted triplets, require high-quality data. A high precision and relatively low recall performance may result in missing some gold labels (i.e. missing links for the knowledge graph). 
%This can be compensated for by increasing the recall. 
However, a high recall and low precision performance (i.e. NoGen achieves the best recall performance in all settings) means that the model returns many results, but most of its predicted labels are incorrect compared to the gold labels. This may introduce much noise (i.e. a noisy dataset) to downstream tasks. We conjecture that the high precision performance of ZS-SKA is due to setting a relatively smaller distance threshold in Sec.~\ref{sec:threshold} for determining the number of relation triplets. In future work, a dynamic threshold could be added to adjust to different datasets.

\subsubsection{Zero-shot Relation Classification}
ZS-SKA also supports the zero-shot relation classification task by providing two name entities in the training set.
Recognizing unseen relations is mainly supported by semantic knowledge augmentation in Sec.~\ref{sec:3.2}.
Therefore, we carry out relation classification experiments on NYT, FewRel and Wiki\_ZSL datasets to better evaluate zero-shot ability of ZS-SKA.

\paragraph{Results on Unbalanced Dataset}%Grammar Done

The experiment results on unbalanced dataset NYT by varying $m$ unseen relations are shown in Table~\ref{tab:mainresultun}.
To make fair comparisons, we use the same splitted NYT dataset and follow the same threshold schema provided by~\cite{10.1145/3459637.3482403}.
We remove the data augmentation module and only implement the prompts generated through the knowledge graph as similar side information in ZSLRC model.
Apparently, the proposed ZS-SKA achieves a substantial gain in precision, recall, and F1-score over other baselines on the NYT dataset.
When the number of unseen relations in the testing set becomes larger, the superiority of ZS-SKA gets more significant and robust. 
Such results indicate the effectiveness of leveraging prompts using virtual labels constructed from the knowledge graph instead of using keywords learned from the distribution of training data on the noisy dataset in ~\cite{10.1145/3459637.3482403}.
\vspace{-2mm}
\paragraph{Results on Balanced Datasets}
\begin{table}[]
\small
\caption{RC results (m=15) on Wiki-ZSL/FewRel.}
\label{tab:mainresult-3}
\centering
\tabcolsep=0.16cm
\begin{tabular}{lccc}
\hline
                & Pre.                 & Rec.                 & F1          \\ \hline
CNN             & 14.58/14.17          & 17.68/20.26          & 15.92/16.67 \\
BiLSTM         & 16.25/16.83          & 18.94/27.62          & 17.49/20.92 \\
BiLSTM\textsubscript{att}     & 16.93/16.48          & 18.54/26.36          & 17.70/20.28 \\
R-BERT          & 17.31/16.95          & 18.82/19.37          & 18.03/18.08 \\
ESIM            & 27.31/29.15          & 29.62/31.59          & 28.42/30.32 \\
CIM             & 29.17/31.83          & 30.58/33.06          & 29.86/32.43 \\
ZS-BERT         & 34.12/35.54          & 34.38/38.19          & 34.25/36.82 \\
NoGen           & \textbf{54.45/66.49} & 29.43/40.05          & 37.56/\textbf{49.38} \\ \hline
\textbf{ZS-SKA} & 41.78/45.03          & \textbf{40.50/51.86} & \textbf{39.30}/46.99 \\ \hline
\end{tabular}
\end{table}

The evaluation results of zero-shot RC on Wiki-ZSL and FewRel are shown in Table~\ref{tab:mainresult-3}.
The results of all baselines are reported by ~\cite{chen2021zsbert, chia-etal-2022-relationprompt} and the result of ZS-SKA is reported by the average of five different random seeds.
For fair comparison, we compare our proposed model with baselines that do not require any training process for additional models such as the generator (GPT-2). 
Obviously, ZS-SKA significantly outperforms other existing models on both balanced datasets for recall value. 
Besides, ZS-SKA has the best F1 performance on Wiki-ZSL.
%Our proposed ZS-SKA outperforms a recently proposed method (ZS-BERT) by 7.66\% precision, 6.12\% recall, and 5.05\% F1-score on Wiki-ZSL, 9.49\% precision, 13.67\% recall, and 10.17\% F1-score on FewRel.
The performance improvement indicates that semantic knowledge augmentation is competitively more beneficial for recognizing unseen relations than only incorporating text description of relations.
%To further compare the robustness of ZS-SKA with the strongest baselines ZS-BERT, NoGen and RelationPrompt, we conduct more experiments with different percentages (varying $m$) of unseen relations in Sec.~\ref{sec:ablation} and Table~\ref{tab:ablation-3}.

\subsection{Analysis}
\subsubsection{Ablation Study}~\label{sec:ablation}
\begin{table}[]
\small
\caption{Ablation study (F1) over ZS-SKA on Wiki-ZSL with different percentages of unseen relations.}
\label{tab:ablation-3}
\centering
\tabcolsep=0.16cm
\begin{tabular}{lccccc}
\hline
                       & 10\% & 20\% & 30\% & 40\% & 50\% \\\hline
ZS-BERT &    58.31  &  19.59   &    17.63 &   11.79  &    9.52  \\
NoGen &    42.72  &  26.20   &    18.20 &   12.28  &    8.69  \\
RelPrompt &    \textbf{67.91}  & \textbf{50.02}   &    \textbf{36.51} &   22.13 &    12.94 \\
Ours\textsubscript{Aug} &    41.44  &  33.57    &    26.00  &   22.04   &    16.00  \\
Ours\textsubscript{Prompts}    &   46.59   &   36.20   &  27.76    &  19.32    &   13.72   \\
Ours\textsubscript{Top2 freq}        &   41.10   & 33.68     & 28.49     &  22.40    &   18.60   \\
Ours\textsubscript{Top5 freq}        &  40.90    &  34.85    & 28.84     &  22.68    &18.25   \\
Ours\textsubscript{ActLabel}&  42.06    & 34.89     &28.85      &22.93      &18.43   \\
Ours\textsubscript{OnlyBert}&   37.35   & 31.80    &  25.80    & 20.75     & 16.51  \\\hline
Ours\textsubscript{All}& 40.93     & 35.99     & 28.97     & \textbf{24.64}     & \textbf{19.27}  \\\hline
\end{tabular}
\end{table}

\begin{table*}[]
\small
\centering
\caption{Examples of sentence generation from seen relations by data augmentation. Words in red are name entities for each sentence. $S(\cdot)$ denotes the super-class of the relation or name entities.}
\label{tab:dataexample}
\begin{tabular}{|c|c|c|c|c|}
\hline
Relation $r$                   & $S(r)$                 & $S(e_1)$               & $S(e_2)$                & Sentence \\ \hline
place\_of\_birth           & location                    & person                    & location                  & \makecell[l]{\textcolor{red}{Jessica} (born in \textcolor{red}{Manchester}) is a British track \\ and field athlete who competes in the heptathlon. }     \\ \hline
place\_of\_death           & location                    & person                    & location                  & \makecell[l]{\textcolor{red}{Johnson} (died in \textcolor{red}{Liverpool}) is a Military track \\ and field athlete who competed in the decathlon. }        \\ \hline
residence                  & location                    & person                    & location                  & \makecell[l]{\textcolor{red}{Mansion} (resided in \textcolor{red}{Villa}) is a Colonial residence \\ and peri alumnus who dominates in the decathlon. }         \\ \hline
country                    & location                    & location                  & location                  & \makecell[l]{\textcolor{red}{Rich} (retired in \textcolor{red}{Arsenal}) is a \textcolor{blue}{European} track and \\ field athlete who competes in the decathlon. }        \\ \hline
educated\_at                    & action                    & person                  & organization                  & \makecell[l]{\textcolor{red}{Jess} (motivate in \textcolor{red}{Liverpool}) is a British aims and \\ professional athlete who educated in the \textcolor{blue}{decathlon}. }        \\ \hline
\end{tabular}
\end{table*}
\vspace{-4mm}
To evaluate the robustness and effectiveness of the zero-shot ability of ZS-SKA, we conduct an ablation study on Wiki-ZSL by removing different modules from ZS-SKA.%, in order to make comparisons with the most SOTA RC models: RelationPrompt and ZS-BERT.
The zero-shot setting is followed by the definition that partial relations are unseen in the testing set~\cite{yinroth2019zeroshot}.
This setting is more competitive because all classes (including both seen and unseen relations) exist in the testing set.
Different with the experiments of specific $m$ values, this is a 113 class classification experiment, including different percentages of unseen relations, which is more related to the real-world scenario.
From Table~\ref{tab:ablation-3}, we observe that ZS-SKA is more robust when increasing the proportions of unseen relations.
The performance drops drastically for ZS-BERT and NoGen when more unseen relations appear.
RelationPrompt is more stable than ZS-BERT and NoGen. 
But the performace also drops a lot starting from 40\% of unseen relations.
%Removing data augmentation module in ZS-SKA performs slightly better than ZS-SKA when 20\% of unseen relations exist.
%Nevertheless, the performance drops more significantly than ZS-SKA.
Though instances generated by data augmentation for unseen relations may include noise, the models with data augmentation can be more robust when large percentages of unseen relations exist in the testing set. 
We also implement models with different ways to construct the prompt such as using top $k$ frequency words, actual label itself to evaluate the virtual label construction in ZS-SKA. Virtual label construction is more effective when 20\% or more of unseen relations exist.
It is because prompts constructed by virtual labels contain the semantic information of unseen relations, which shortens the distance between the query sentence of an unseen relation with the unseen relation prototype.

\subsubsection{Case Study}%Grammar Done
\begin{figure}[!htb]
\centering
{
\includegraphics[width=3cm]{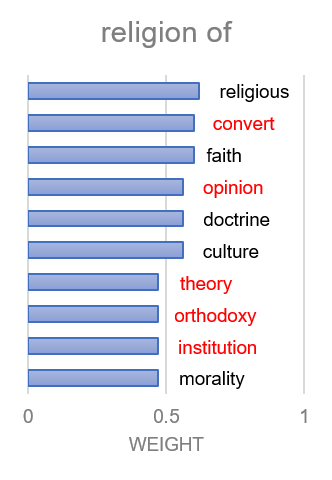}
%\caption{fig1}
}
\hspace{-2mm}
{
\includegraphics[width=3cm]{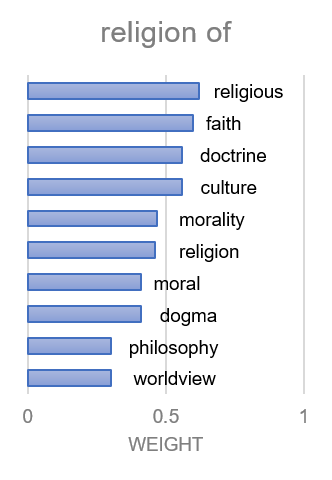}
}
\caption{Denoising in virtual label construction.}
\label{fig:prompt} 
\end{figure}

\paragraph{Virtual Label Construction}
Figure~\ref{fig:prompt} shows an example of ranking the top ten components of the constructed virtual label before denoising and after denoising.
The virtual labels shown in Figure~\ref{fig:prompt} are generated by Algorithm~\ref{alg:prompts}.
The red words are irrelevant to the relation `religion\_of'.
After we refine the virtual label sets using the distance metric, these irrelevant words are filtered out in our virtual label sets, removing the noise in the knowledge graph.

\begin{figure}[htp] 
 \center{\includegraphics[height=2.5cm,width=7.5cm]{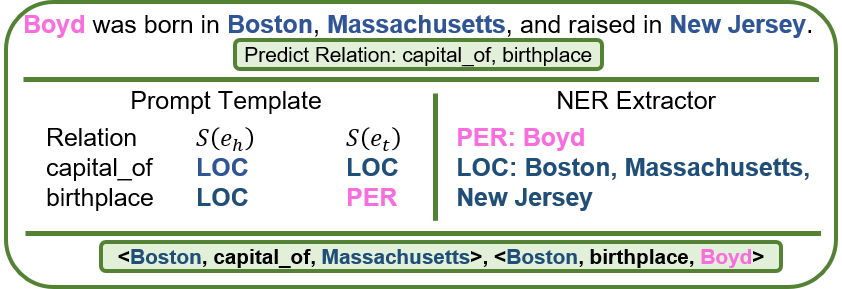}}
 \caption{\label{fig:ner} Example of using Name Entity Extractor to extract relation triplets.} 
 \end{figure}

\paragraph{Name Entity Extractor}
Figure~\ref{fig:ner} shows an example of how ZS-SKA uses the name entity extractor with super-classes information to extract relation triplets.
ZS-SKA includes two steps for RTE. 
First, unseen relations are predicted by semantic knowledge augmentation.
Based on the predicted relations, super-class information such as `LOC' and `PER' can be accessed from the prompt template.
Second, the NER extractor is implemented to extract the types of name entities.
For example, the relation `birthplace' happens between `PER' and `LOC' according to the template.
The filter in the NER extractor selects `Boyd' in `PER' and `Boston' in `LOC'.
Similarly, relation `capital\_of' only happens between `LOC' and `LOC', so the filter in the NER extractor selects `Boston' and `Massachusetts' in `LOC'.
Note that all locations in `LOC' in Figure~\ref{fig:ner} are ranked based on the possibility score. 
Then, the predicted relations and entities extracted by the NER extractor construct the final relation triplets.

\vspace{-2mm}
\paragraph{Data Augmentation}%Done
Table~\ref{tab:dataexample} shows an example of the augmented data following the translating rule on the Wiki-ZSL dataset.
The relation `place\_of\_birth' is a seen class, and the other four relations are from unseen classes.
We use data augmentation to generate augmented training instances for these unseen relations.
We observe that if the super-class of both the relation and two name entities are the same, the generated sentences have a good quality with the name entities having unseen relations.
If the super-class of the relation or two name entities of unseen relation is different from that of the seen relation,  though the generated sentences contain the tone of the unseen relation (words in blue), the original two name entities do not have the target unseen relation.
For example, the generated sentence of relation 'country' can be explained that Arsenal is from a European country, but such relation is lost between the two name entities `Rich' and `Arsenal'.
Therefore, we follow the rule of using the relation and name entities from the same super-class with that of unseen relations to generate high-quality augmented instances for training in ZSL.

\subsubsection{Hyperparameter Sensitivity}%Grammar Done
\begin{figure}[!htbp]
\centering
{
\includegraphics[width=3.8cm]{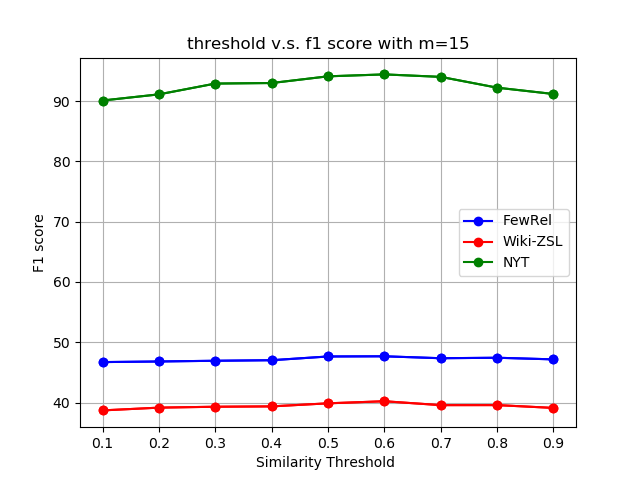}
%\caption{fig1}
}
\hspace{-6mm}
{
\includegraphics[width=3.8cm]{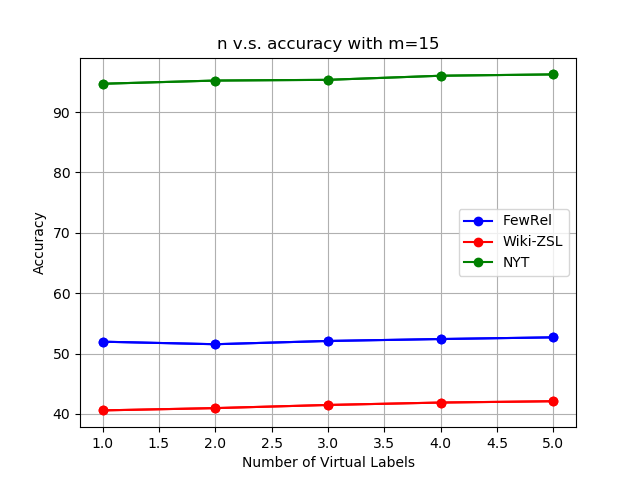}
}
\caption{Effects on varying threshold $\tau$ and number of virtual labels $n$ on three datasets.}
\label{fig:effects} 
\end{figure}

We examine how some hyperparameters, including threshold $\tau_{s}$ for denoising virtual label sets and the number of virtual labels $n$, affect ZS-SKA's performance.
By fixing $m$ = 15 and varying $\tau_{s}$ and $n$, the results are exhibited in Figure~\ref{fig:effects}.
We find that $\tau_{s}$ and $n$ affect the noisy dataset more than the balanced dataset.
We think that because both $\tau_{s}$ and $n$ are used for removing noise and getting more related semantic information in prompts, the noise in prompts may impact more on noisy datasets because they are more sensitive to the noise.

It is obvious that $\tau_{s}$ does have an impact on the performance.
If the threshold $\tau_{s}$ is between 0.5 and 0.6, it achieves the best performance on all three public datasets.
This is reasonable that when $\tau_{s}$ is too low, most connected nodes in the knowledge graph are used to construct virtual label words, and more noise may be obtained from the relation.
%Thus, when building the prompts for each relation, it is more likely to bring the noise to the relation.
In contrast, when $\tau_{s}$ gets too high, some highly related nodes are filtered out to construct virtual labels.
%We would suggest setting $\tau_{s}$ between 0.5 to 0.6 to derive satisfying results across datasets.
We also find that increasing the number of related words $n$ to construct virtual labels can achieve better performance.
It is reasonable because, including more nodes (words) from the knowledge graph to construct the virtual label representing the relation, more semantic information is contained, leading to a shorter distance.% between the query sentence embedding with the prototype constructed from the prompts.

\section{Conclusion and Future Work}%Grammar Done
We propose a ZS-SKA utilizing semantic knowledge augmentation to extract unseen relation triplets with no labeled data available for training to tackle with zero-shot RTE.
The experiments show that with augmented instances, prompts generated through a knowledge graph, and a NER extractor with prompts, ZS-SKA outperforms other SOTA zero-shot RTE models.
We have also conducted extensive experiments to study different aspects of ZS-SKA, from ablation study, and case study to hyperparameter sensitivity, and demonstrate the effectiveness and robustness of our proposed model.
In future work, we plan to explore: (1) Different ways of instance generation and prompt designs for semantic augmented data.
(2) Better approaches for constructing virtual labels in the prompt template.
(3) More SOTA data augmentation techniques to generate data for zero-shot tasks to further improve the performance.

\nocite{*}
\section{Bibliographical References}\label{sec:reference}

\bibliographystyle{lrec-coling2024-natbib}
\bibliography{lrec-coling2024-example}

\section{Appendix}
\label{sec:appendix}

\subsection{Dataset Description}
\label{sec:app_data}
\begin{table}[!htb]
\caption{The statistics of each dataset.}
\centering
\label{tab:dataset}
\begin{tabular}{cccc}
\hline
         & \#instances & \#relations & avg. Len. \\ \hline
FewRel   & 56,000      & 80          & 24.95     \\
Wiki-ZSL & 94,383      & 113         & 24.85     \\
NYT      & 134,152     & 53          & 38.81          \\ \hline
\end{tabular}
\end{table}

\begin{itemize}
\item \textbf{\textit{FewRel~\cite{han-etal-2018-fewrel}.}}
The FewRel dataset is a human-annotated balanced few-shot RC dataset consisting of 80 types of relations, each of which has 700 instances.
\item \textbf{\textit{Wiki-ZSL~\cite{chen2021zsbert}.}}
The Wiki-ZSL dataset is a subset of Wiki-KB~\cite{sorokin-gurevych-2017-context}, which filters out both the 'none' relation and relations that appear fewer than 300 times.
\item \textbf{\textit{NYT~\cite{10.1007/978-3-642-15939-8_10}.}}
The NYT dataset was generated by aligning Freebase relations with the New York Times.
There are 53 possible relations in total.
It is an unbalanced noisy dataset.
\end{itemize}

\subsection{Parameter Settings}~\label{sec:app_para}
\begin{table}[!htb]
\caption{Parameter Settings}
\label{tab:parameter-3}
\centering
\begin{tabular}{lr}
\hline
Parameter                            & Value     \\ \hline
Word Embedding Dimension     & 768        \\
Hidden Layer Dimension        & 300       \\
Sentence Max Length             & 128\\
Convolutional Window Size        & 3         \\
Batch Size                           & 4         \\
Initial Learning Rate $\alpha$       & 0.01      \\
Weight Decay                     &$10^{-5}$\\
Number of Hops $K$                        & 1    \\
Similarity Threshold $\tau_{s}$                      & 0.6 \\
Distance Threshold $\tau_{d}$   & 0.05 \\
NER Threshold $\tau_{e}$  & 0.5 \\
Number of Virtual Label $n$    & 5\\
\hline
\end{tabular}
\end{table}

For the hyperparameter and configuration of ZS-SKA, we implement ZS-SKA with PyTorch and optimize it with an SGD optimizer.
The initial learning rate is selected via the grid search within
the range of $\begin{Bmatrix}
1e-1, 1e-2, 1e-3, 1e-4
\end{Bmatrix}$ for minimizing the loss, the cosine similarity threshold is selected from 0 to 1 with step size 0.1.
The distance threshold for determining the number of triplets in a given sentence is set to 0.05, which is explored in the validation set.
NER threshold is selected from 0.1 to 0.9 with a step size of 0.2.
%The model for testing is selected based on 
Table~\ref{tab:parameter-3} shows other parameters.
We follow the early stopping strategy when selecting the model for testing. The model is evaluated on the validation set every 50 epochs. 
The time for training is around 6 hours depending on the computing resources. GPU with 16G memory is required for training.

\subsection{Limitations}
Given the progress made to date with the work we propose in this paper, we view the following current limitations as some opportunities to build on in future work. 
First, data augmentation is based on word-level transformation. With the development of generation models, more state-of-the-art data augmentation techniques can be implemented to generate data for zero-shot tasks to further improve the performance. Second, the proposed prompt method depends on information from a fixed knowledge graph, which means it can not deal with the scenario if the unseen label is an out-of-vocabulary word. We have not considered this scenario because all classes from the three public datasets are well-known words or phrases.
In future work, to get prompt information when the class word does not exist in the knowledge graph, we will consider directly using label descriptions or text generation models such as GPT-2 to generate label explanations.

\subsection{Ethical Considerations}

\paragraph{Data Bias} Our proposed model ZS-SKA is specifically intended for zero-shot relation triplet extraction or zero-shot relation classification tasks. We perform experiments on three public datasets. However, the model's performance may be subject to bias when applied to other datasets with significantly different distributions or in new domains.
Therefore, we advise exercising caution when assessing the generalizability and fairness of the model.

\paragraph{Computing Cost} Our model needs the use of GPU training, which imposes a computational burden. We acknowledge that this burden has an adverse environmental impact on carbon emissions. Specifically, our research requires 6 hours of training on a single GPU card for each task. In total, we have 6 $\times$ 5 (each task has 5 runs) $\times$ 15 tasks (6 tasks in RTE, 4 tasks in RC, and 5 tasks in ablation study) = 450 hours of training, resulting in 112.5lbs of carbon dioxide.

\end{document}